\documentclass[runningheads]{llncs}
\usepackage[hidelinks, colorlinks=true, linkcolor=blue, citecolor=blue, breaklinks=true]{hyperref}
\usepackage[T1]{fontenc}
\usepackage{adjustbox}
\usepackage{graphicx}
\usepackage{multirow}
\usepackage{xcolor}
\usepackage{colortbl}
\usepackage{booktabs}
\usepackage{ulem}
\usepackage{pifont}
\usepackage{amsfonts,amssymb,amsmath}
\usepackage{dsfont}
\usepackage{float}
\usepackage{subfigure}
\usepackage{subcaption}
\usepackage{marvosym}
\usepackage{wasysym}

\usepackage{bbding}
\usepackage{fancyhdr}
\usepackage{bm}       
\usepackage{booktabs}
\usepackage{multirow}
\usepackage{pdflscape}
\usepackage{pifont}
\usepackage[utf8]{inputenc}
\definecolor{lightpink1}{HTML}{FFD5D9}
\definecolor{lightpink2}{HTML}{FFE2E6}
\definecolor{lightblue1}{HTML}{D6ECFF}  
\definecolor{lightblue2}{HTML}{E9F4FF}  
\definecolor{lightpink3}{HTML}{FFD0DA}
\definecolor{lightpink4}{HTML}{FBEEE2}
\definecolor{lightpink5}{HTML}{FBF2E3}
\definecolor{lightW1}{HTML}{F5F5F5}
\definecolor{lightW2}{HTML}{F8F4ED}
\definecolor{lightW3}{HTML}{F7F4ED}
\definecolor{lightW4}{HTML}{F8F8F8}
\definecolor{lightW5}{HTML}{EEE6DC}
\newcommand{\gain}[1]{\textcolor{gray}{\scriptsize$\uparrow#1$}}
\newcommand{\loss}[1]{\textcolor{gray}{\scriptsize$\downarrow#1$}}

\newcommand{\ourscell}{}

\begin{document}

\title{SAIF: A Stability-Aware Inference Framework for Medical Image Segmentation with Segment Anything Model}

\author{
Ke~Wu\inst{1}\and
Shiqi~Chen\inst{1} \and
Yiheng~Zhong\textsuperscript{\dag}\inst{1}\and
Hengxian~Liu\inst{1} \and
Yingxue~Su\inst{1} \and
Yifang~Wang\inst{2} \and
Junhao~Jin\inst{3} \and
Guangyu~Ren\textsuperscript{\dag}\inst{1} \
}
\institute{Xi'an Jiaotong-Liverpool University, Suzhou, China \and University College London, London, United Kingdom \and Anhui Jianzhu University, Hefei, China  \\ \email{JohnZhongJohn@outlook.com}} 
\maketitle

\begingroup

\renewcommand\thefootnote{}\footnotetext{\textsuperscript{\dag}Corresponding author.}
\endgroup

\begin{abstract}
Segment Anything Model (SAM) enable scalable medical image segmentation but suffer from inference-time instability when deployed as a frozen backbone. In practice, bounding-box prompts often contain localization errors, and fixed threshold binarization introduces additional decision uncertainty. These factors jointly cause high prediction variance, especially near object boundaries, degrading reliability. We propose the Stability-Aware Inference Framework (SAIF), a training-free and plug-and-play inference framework that improves robustness by explicitly modeling prompt and threshold uncertainty. SAIF constructs a joint uncertainty space via structured box perturbations and threshold variations, evaluates each hypothesis using decision stability and boundary consistency, and introduces a stability-consistency score to filter unstable candidates and perform stability-weighted fusion in probability space. Experiments on Synapse, CVC-ClinicDB, Kvasir-SEG, and CVC-300 demonstrate that SAIF consistently improves segmentation accuracy and robustness, achieving state-of-the-art performance without retraining or architectural modification. Our anonymous code is released at \url{https://anonymous.4open.science/r/SAIF}.


\end{abstract}

\section{Introduction}
\begin{figure}[t]
    \centering
    \includegraphics[width=\linewidth]{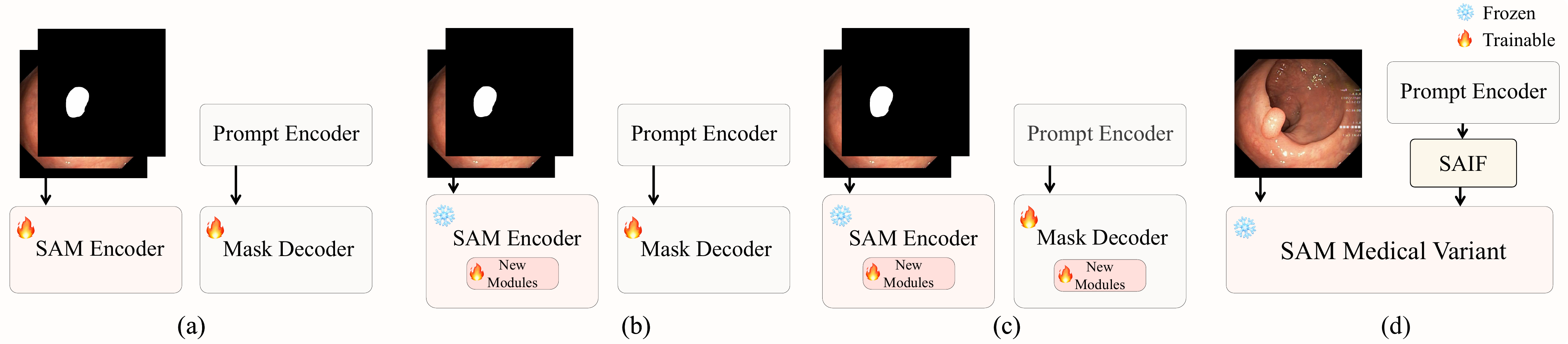} 
    \caption{Overview of medical adaptations of the Segment Anything Model. (a) Full retraining on large-scale medical data, (b) parameter-efficient fine-tuning, (c) structural modifications to the encoder and mask decoder, and (d) inference-time operating conditions affecting output reliability, addressed by approach.}
    \label{fig:fig1_overview}
\end{figure}

Medical image segmentation is a fundamental task in medical image analysis, providing precise delineation of organs and lesions to support diagnosis, treatment planning, and longitudinal follow-up~\cite{survey}. Although SAM~\cite{SAM2023} offers a promising promptable segmentation paradigm, direct transfer to medical images is often unstable under domain shifts. To improve medical applicability, existing adaptations of SAM fall into two lines shown in Fig.~\ref{fig:fig1_overview}. The first improves medical alignment via full retraining or parameter-efficient fine-tuning on large-scale medical data (Fig.~\ref{fig:fig1_overview}(a,b))~\cite{MedSAM2023,SAMed2024,SAMAdapter2023ICCVW}. The second enhances fine-grained anatomical modeling by modifying key components such as the encoder and mask decoder (Fig.~\ref{fig:fig1_overview}(c))~\cite{PGSAM2025,HSAM2024,Tang2024DACL}. Despite substantial progress, these gains stem from training or architectural changes and do not directly address deployment settings where the backbone is frozen. In such settings, output reliability is governed by inference-time operating conditions, prompt quality and binarization rules~\cite{SAMUS2024}, yet systematic modeling and control of the resulting instability remain limited.

Accordingly, we focus on inference-time factors in pipelines (Fig.~\ref{fig:fig1_overview}(d)). Bounding-box prompts are widely used due to annotation efficiency and integration with detectors for inference and pseudo-label generation~\cite{SAM2023}. However, boxes from detectors or rapid interaction contain localization and scale errors and SAM outputs are typically binarized with fixed rules such as thresholding, introducing uncertainty. Prompt noise and threshold variation make predictions sensitive to small box perturbations and threshold changes, leading to run-to-run variance that is amplified near object boundaries and degrades reproducibility~\cite{Huang2024MedSAMEval}.
 
Motivated by this gap, we propose SAIF, a plug-and-play, training-free inference framework that improves reliability for box-prompted medical segmentation with frozen SAM backbones. SAIF is built on a simple verifiable principle: a reliable prediction should remain consistent under small, reasonable perturbations of the bounding box and threshold. Accordingly, SAIF enumerates candidate hypotheses over a joint box, thresholding uncertainty space and evaluates each via two complementary criteria: decision stability, measured by soft IoU against a reference mask; and boundary consistency, measured by boundary-pixel displacement under jitter. Unstable hypotheses are filtered out, and the remaining candidates are fused via stability-weighted probability aggregation to yield consistent, reliable outputs. Our main contributions are as follows:

\begin{enumerate}
\item[(i)] We propose SAIF, a training-free plug-and-play inference framework that performs stability-driven hypothesis screening and fusion under noisy box prompts and fixed threshold decision rules, substantially improving medical SAM reliability without retraining or architectural modification.

\item[(ii)] We formulate box prompting and threshold selection as a joint uncertainty space, proposing a stability-consistency (SC) score for decision stability and
boundary consistency to guide hypothesis selection and probability fusion.

\item[(iii)] We validate SAIF on four benchmarks. Results show consistent state-of-the-art improvements across diverse anatomical structures and cross-distribution settings, and in certain tasks SAIF surpasses fully supervised methods.
\end{enumerate}

\section{Methodology} 

\subsection{Overview of the Proposed Framework}
\begin{figure}[t]
    \centering
    \includegraphics[width=\linewidth]{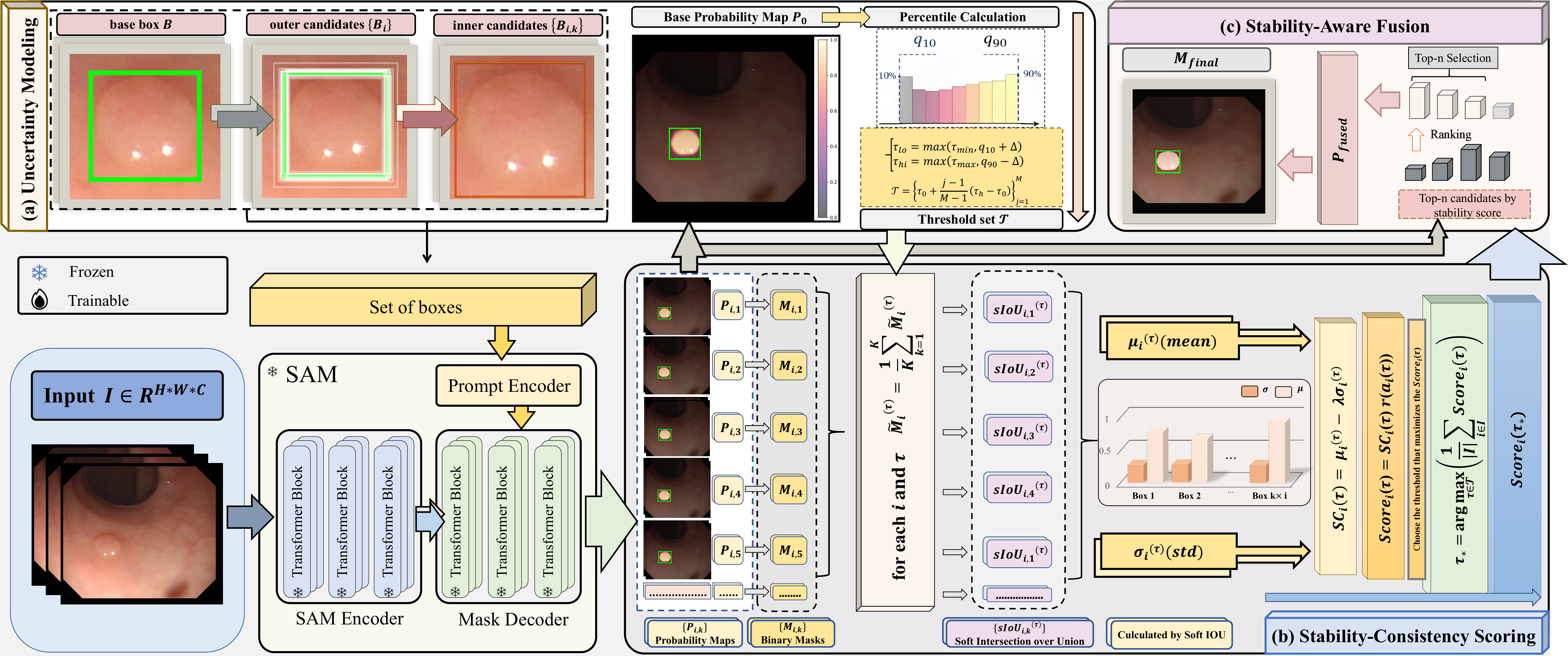} 
    \caption{This diagram shows the SAIF framework for stability-aware medical image segmentation. It starts with perturbing the original box prompt into multiple outer and inner candidates, which are processed through the model to generate probability maps. Stability-consistency scoring ranks candidates based on Soft IoU and percentiles. The top-$n$ candidates are selected and fused to produce the final segmentation output, using no additional network execution or training.}
    \label{fig:fig2_overview}
\end{figure}

We propose a training-free, stability-aware inference framework for box-prompted medical image segmentation. Inference is modeled as robustness over a bounded uncertainty set induced by box perturbations and image-adaptive thresholds. A prediction is reliable if its binary output remains stable across this set. Using cached probability maps, we select an image-wise shared threshold $\tau_\ast$ via stability maximization and fuse the top candidates for the final mask, without additional network execution. The process is shown in Figure~\ref{fig:fig2_overview}.

\subsection{Uncertainty Modeling}
\label{sec:uncertainty_modeling}
Uncertainty arises from imprecise box prompts and threshold-based binarization. We therefore model inference as robustness over the resulting joint uncertainty set, consisting of a finite prompt family and an image-adaptive threshold set.

Let \(I\in\mathbb{R}^{H\times W\times C}\) be the input and \(B=(x_1,y_1,x_2,y_2)\) the provided box. We define a two-level coarse-to-fine prompt family:
\begin{equation}
\begin{aligned}
B_i &= \mathrm{Jitter}\!\left(\mathrm{Scale}(B,\alpha_i),\, \delta_{\mathrm{out}}\right), \quad i=1,\dots,N,\ \alpha_i\in\mathcal{S},\\
B_{i,k} &= \mathrm{Jitter}(B_i,\, \delta_{\mathrm{in}}), \quad k=2,\dots,K,\ \text{with } B_{i,1} = B_i,\ \delta_{\mathrm{in}}<\delta_{\mathrm{out}}.
\end{aligned}
\end{equation}

Let \(\mathcal{S}\) be a small discrete set centered at \(1\), where \(\alpha<1\) tightens and \(\alpha>1\) loosens the box. We refer to \(\{B_i\}_{i=1}^{N}\) as outer candidates and \(\{B_{i,k}\}_{k=1}^{K}\) as inner jitters, where the former provides coarse-scale hypotheses and the latter probes fine local stability around each \(B_i\). For each \(i\), we set \(B_{i,1}=B_i\) and draw \(K-1\) additional jitters. The operator \(\mathrm{Jitter}(B;\delta)\) perturbs box boundaries by size-relative offsets sampled from \([-\delta,\delta]\). All candidates are clamped to the image domain; invalid boxes are discarded, yielding at most \(NK\) valid prompts and thus at most \(NK\) forward passes per image.

For each valid prompt, we cache \(P_{i,k}=f(I,B_{i,k})\). The original prediction \(P_0=f(I,B)\) serves as the reference for threshold construction. Let \(\mathcal{I}\) denote valid outer candidates. From \(P_0\), we form an image-specific threshold set \(\mathcal{T}\): let \(q_{10}\) and \(q_{90}\) be the 10th and 90th percentiles of \(P_0\) within the clamped box region, and define a robust interval \([\tau_{\mathrm{lo}},\tau_{\mathrm{hi}}]\) to avoid tail-induced degenerate masks:
\begin{equation}
\mathcal{T}=
\left\{
\tau_{\mathrm{lo}}+\frac{j-1}{M-1}\left(\tau_{\mathrm{hi}}-\tau_{\mathrm{lo}}\right)
\right\}_{j=1}^{M},
\ \text{s.t. }\ 
\begin{cases}
\tau_{\mathrm{lo}}=\max(\tau_{\min},\, q_{10}+\Delta),\\
\tau_{\mathrm{hi}}=\min(\tau_{\max},\, q_{90}-\Delta),
\end{cases}
\end{equation}
where $[\tau_{\min},\tau_{\max}]$ are global conservative bounds, $\Delta$ is a safety margin, and $M$ controls the discretization density. If $\tau_{\mathrm{lo}}\ge\tau_{\mathrm{hi}}$, we set $\mathcal{T}=\{\mathrm{clip}(\mathrm{median}(P_0),\tau_{\min},\tau_{\max})\}$. This focuses subsequent stability scoring on informative operating regions.

\subsection{Stability-Consistency Scoring}
\label{sec:stability_scoring}
Given cached predictions $\{P_{i,k}\}$ and the threshold set $\mathcal{T}$, we score stability in the joint prompt--threshold space by binarizing each map at $\tau\in\mathcal{T}$,
$M_{i,k}^{(\tau)}=\mathbb{I}(P_{i,k}>\tau)$.
For each outer candidate $B_i$, we form a pixel-wise consensus
$\tilde{M}_i^{(\tau)}=\frac{1}{K}\sum_{k=1}^{K} M_{i,k}^{(\tau)}$
and measure agreement via soft IoU:
\begin{equation}
\mathrm{sIoU}_{i,k}^{(\tau)} =
\frac{\sum_{x,y} \min\!\left(M_{i,k}^{(\tau)}(x,y),\, \tilde{M}_i^{(\tau)}(x,y)\right)}
{\sum_{x,y} \max\!\left(M_{i,k}^{(\tau)}(x,y),\, \tilde{M}_i^{(\tau)}(x,y)\right) + \epsilon},
\end{equation}
where \(\epsilon>0\) prevents division by zero. We evaluate stability after binarization to match the final decision rule and to be robust to calibration shifts that do not change the induced masks.
For each outer candidate $i$ and threshold $\tau\in\mathcal{T}$, we summarize
$\{\mathrm{sIoU}_{i,k}^{(\tau)}\}_{k=1}^{K}$ over $k$ by the mean $\mu_i^{(\tau)}$ and standard deviation $\sigma_i^{(\tau)}$. We then define a variance-penalized stability-consistency score with area regularization:
\begin{equation}
\mathrm{SC}_i(\tau)=\mu_i^{(\tau)}-\lambda\,\sigma_i^{(\tau)}, \qquad
\mathrm{Score}_i(\tau)=\mathrm{SC}_i(\tau)\, r\!\left(a_i(\tau)\right),
\end{equation}
where $\lambda\ge 0$ weights the variability penalty and $a_i(\tau)$ is the mean foreground occupancy across the $K$ masks. The area gate $r(\cdot)$ down-weights degenerate solutions that produce nearly-empty or nearly-full masks:
\begin{equation}
r(a)=
\begin{cases}
1, & a_{\min} < a < a_{\max},\\
\gamma, & \text{otherwise},
\end{cases}
\end{equation}
with loose bounds $a_{\min}$ and $a_{\max}$ reflecting plausible target size, and $\gamma<1$. We select an image-wise shared threshold by maximizing the average candidate score over $\mathcal{T}$, enforcing a single decision rule shared across all perturbed prompts without additional training. When multiple thresholds attain the same maximum, we choose the smaller $\tau$ to obtain a more conservative decision boundary.

\begin{equation}
\tau_{\ast} = \arg\max_{\tau \in \mathcal{T}}
\left( \frac{1}{|\mathcal{I}|} \sum_{i\in\mathcal{I}} \mathrm{Score}_i(\tau) \right).
\end{equation}

\subsection{Stability-Aware Fusion}
\label{sec:inference_fusion}
Given the threshold $\tau_{\ast}$ from Section~\ref{sec:stability_scoring}, the final segmentation is obtained  via selection and score-weighted fusion over the cached predictions $\{P_{i,k}\}$, requiring no network execution beyond the $NK$ forward passes used to compute them. We select the top-$n$ outer candidates by their stability score $\mathrm{Score}_i(\tau_{\ast})$:
\begin{equation}
\mathcal{I}_{\mathrm{sel}}
=
\operatorname*{arg\,top\text{-}n}_{i\in\mathcal{I}}
\ \mathrm{Score}_i(\tau_{\ast}).
\end{equation}

Next, for each selected candidate, we average its $K$ inner-jitter predictions to reduce sensitivity to small prompt perturbations while preserving the candidate-specific consensus,
$\bar{P}_i=\frac{1}{K}\sum_{k=1}^{K}P_{i,k}$. A score-normalized convex combination fuses selected candidates, stressing stable hypotheses and merging ones:

\begin{equation}
w_i=\frac{\mathrm{Score}_i(\tau_{\ast})}{\sum_{j\in\mathcal{I}_{\mathrm{sel}}}\mathrm{Score}_j(\tau_{\ast})},
\qquad
P_{\mathrm{fused}}=\sum_{i\in\mathcal{I}_{\mathrm{sel}}} w_i\,\bar{P}_i.
\end{equation}

Finally, we apply the shared threshold to produce the binary mask in the same decision space used for stability scoring:
\begin{equation}
M_{\mathrm{final}}=\mathbb{I}\!\left(P_{\mathrm{fused}}>\tau_{\ast}\right).
\end{equation}

This inference acts as a reliability filter that explicitly selects, stabilizes, and fuses prompt hypotheses based on stability-derived scores.

\section{Experiment} \label{sec:exp}
\subsection{Experimental Setup}

\noindent\textbf{Dataset.} We evaluate our training-free method on the MICCAI 2015 Synapse multi-organ CT dataset~\cite{Synapse2015}, which contains 3,779 contrast-enhanced abdominal CT slices. Following SAMed~\cite{SAMed2024}, we resize all slices to 224$\times$224 resolution. The evaluation covers eight organs: aorta, gallbladder, spleen, left kidney, right kidney, liver, pancreas, and stomach. We also evaluate on three widely used polyp segmentation benchmark datasets: CVC-ClinicDB~\cite{CVCClinicDB2015} (612 colonoscopy images, 288$\times$384), Kvasir-SEG~\cite{KvasirSEG2020} (1000 image-mask pairs with varied resolutions, collected by Vestre Viken Health Trust, Norway), and CVC-300~\cite{Vazquez2017} (60 images, 500$\times$574, each depicting a single polyp).

\noindent\textbf{Implementation Details.} For fair comparison, all experiments are conducted on the same test sets as prior work, implemented in PyTorch on an RTX A2000 GPU with fixed random seeds. The proposed method operates entirely at inference time and introduces no additional learnable parameters. For all SAM variants, we strictly follow the original implementation settings, integrate our modules, and perform inference using the provided pretrained models.
Unless otherwise specified, we use \(\mathcal{S}=\{0.9,\,1.0,\,1.1\}\), \(K=8\), \(M=7\), \(\lambda=0.3\), \(\mathrm{topn}=3\), and \(\epsilon=10^{-6}\). For box perturbations, we use \(\delta_{\mathrm{out}}=0.04\) and \(\delta_{\mathrm{in}}=0.01\).

\subsection{Comparisons with State-of-the-art Methods}

\newcommand{\DiffUp}{\raisebox{0.08ex}{\scalebox{0.85}{$\blacktriangle$}}}   
\newcommand{\DiffDown}{\raisebox{0.00ex}{\scalebox{0.85}{$\blacktriangledown$}}} 

\renewcommand{\gain}[1]{\textcolor{gray}{\DiffUp\kern0.20em\scalebox{0.92}{#1}}}
\renewcommand{\loss}[1]{\textcolor{gray}{\DiffDown\kern0.20em\scalebox{0.92}{#1}}}

\begin{table*}[!t]
\centering
\small
\caption{Quantitative comparison on polyp segmentation benchmarks. All SAM-based methods use dataset-provided box prompts, while train-based methods are prompt-free. Light gray cells indicate methods equipped with the proposed strategy. Red values denote best-performing results, and purple values indicate improvements larger than 5\% over the corresponding baselines. mDice: the mean Dice coefficient; mIoU: the mean Intersection-over-Union.}
\label{tab:polyp_bench}
\setlength{\tabcolsep}{8pt}
\renewcommand{\arraystretch}{1.18}
\resizebox{1.0\textwidth}{!}{
\begin{tabular}{c|l|cc|cc|cc}
\toprule

\multirow[c]{2}{*}{\textbf{Category}}
& \multirow[c]{2}{*}{\textbf{Methods}}
& \multicolumn{2}{c|}{\cellcolor{lightgray}\textbf{CVC-ClinicDB}}
& \multicolumn{2}{c|}{\cellcolor{lightgray}\textbf{Kvasir-SEG}}
& \multicolumn{2}{c}{\cellcolor{lightgray}\textbf{CVC-300}} \\

& & \cellcolor{lightpink2}\textbf{mDice}$\uparrow$ & \cellcolor{lightpink1}\textbf{mIoU}$\uparrow$
& \cellcolor{lightpink2}\textbf{mDice}$\uparrow$ & \cellcolor{lightpink1}\textbf{mIoU}$\uparrow$
& \cellcolor{lightpink2}\textbf{mDice}$\uparrow$ & \cellcolor{lightpink1}\textbf{mIoU}$\uparrow$ \\
\midrule

\multirow{8}{*}{\rotatebox[origin=c]{90}{\centering\bfseries\shortstack{Train\\based}}}
& U-Net~\cite{UNet2015}        & 82.31 & 75.56 & 81.84 & 74.63& 71.07 & 62.76 \\
& U-Net++~\cite{UNetPlus2018}        & 79.45 & 72.92 & 82.19 & 74.35 & 70.72 & 62.46\\
& TransUNet~\cite{TransUNet2021}      & 89.46 & 85.02 & 86.91 & 80.58 & - & - \\
& EU-Net~\cite{EUNet2021}       & 90.24 & 84.67 & 90.81 & 85.42 & 83.76 & 76.55 \\
& PraNet~\cite{PraNet2020}       & 90.20 & 85.82 & 90.15 & 84.83& 87.32 & 80.48 \\
& SwinUNet~\cite{SwinUNet2021}       & 90.53 &  85.52 & 88.92 & 82.06 & - & - \\
& MSNet~\cite{MSNet2021}       & 92.19 & 87.98 & 90.75 & 86.22 & 86.93 & 80.77 \\
& Polyp-PVT~\cite{PolypPVT2022}       & \textcolor{red}{93.72} & \textcolor{red}{88.97} & 91.70 & 86.42 & 90.03 & 83.35 \\
\midrule

\multirow{7.5}{*}{\rotatebox[origin=c]{90}{\centering\bfseries\shortstack{SAM-based\\Methods}}}
& SAM~\cite{SAM2023}            & 33.29 & 25.64 & 61.48 & 53.74 & 45.00 & 38.62 \\
& MedSAM~\cite{MedSAM2023}         & 75.25& 63.17 & 79.21 & 67.74 & 77.78 & 64.99\\

& \cellcolor{gray!20}\textbf{MedSAM + Ours}
  & \cellcolor{gray!20}\textbf{83.33}\ourscell\,\gain{\textcolor{purple}{8.08}}
  & \cellcolor{gray!20}\textbf{71.99}\ourscell\,\gain{\textcolor{purple}{8.82}}
  & \cellcolor{gray!20}\textbf{85.16}\ourscell\,\gain{\textcolor{purple}{5.95}}
  & \cellcolor{gray!20}\textbf{76.35}\ourscell\,\gain{\textcolor{purple}{8.61}}
  & \cellcolor{gray!20}\textbf{81.97}\ourscell\,\gain{4.19}
  & \cellcolor{gray!20}\textbf{70.91}\ourscell\,\gain{\textcolor{purple}{5.92}} \\

& SAM Adapter~\cite{SAMAdapter2023ICCVW}    & 85.54  & 78.54 & 86.31 & 80.25 & 86.19 & 81.77 \\

& \cellcolor{gray!20}\textbf{SAM Adapter + Ours}
  & \cellcolor{gray!20}\textbf{92.17}\ourscell\,\gain{\textcolor{purple}{6.63}}
  & \cellcolor{gray!20}\textbf{85.22}\ourscell\,\gain{\textcolor{purple}{6.68}}
  & \cellcolor{gray!20}\textbf{91.33}\ourscell\,\gain{\textcolor{purple}{5.02}}
  & \cellcolor{gray!20}\textbf{84.92}\ourscell\,\gain{4.67}
  & \cellcolor{gray!20}\textbf{92.33}\ourscell\,\gain{\textcolor{purple}{6.14}}
  & \cellcolor{gray!20}\textbf{85.50}\ourscell\,\gain{3.73} \\

& SAMed~\cite{SAMed2024} & 83.34 & 76.32 & 88.01 & 81.61 & 89.23 & 83.32 \\

& \cellcolor{gray!20}\textbf{SAMed + Ours}
  & \cellcolor{gray!20}\textbf{87.77}\ourscell\,\gain{4.43}
  & \cellcolor{gray!20}\textbf{82.32}\ourscell\,\gain{\textcolor{purple}{6.00}}
  & \cellcolor{gray!20}\textbf{\textcolor{red}{92.96}}\ourscell\,\gain{4.95}
  & \cellcolor{gray!20}\textbf{\textcolor{red}{89.42}}\ourscell\,\gain{\textcolor{purple}{7.81}}
  & \cellcolor{gray!20}\textbf{\textcolor{red}{94.75}}\ourscell\,\gain{\textcolor{purple}{5.52}}
  & \cellcolor{gray!20}\textbf{\textcolor{red}{90.54}}\ourscell\,\gain{\textcolor{purple}{7.22}} \\

\bottomrule
\end{tabular}
}
\end{table*}

As shown in Table~\ref{tab:polyp_bench}, the proposed inference strategy yields consistent and significant gains in both mDice and mIoU across CVC-ClinicDB, Kvasir-SEG, and CVC-300 for all SAM-based medical adaptation models. Compared with their original counterparts, SAM variants achieve average improvements of approximately 4--8\% in mDice and 5--9\% in mIoU. For instance, MedSAM + Ours improves mDice by $\uparrow$8.08\%, $\uparrow$5.95\%, and $\uparrow$4.19\% on the three benchmarks, with corresponding mIoU gains of $\uparrow$8.82\%, $\uparrow$8.61\%, and $\uparrow$5.92\%. With the proposed strategy, SAM Adapter achieves mDice and mIoU scores of 92.17\% and 85.22\% on CVC-ClinicDB, 91.33\% and 84.92\% on Kvasir-SEG, and 92.33\% and 85.50\% on CVC-300, exceeding representative fully supervised approaches such as Polyp-PVT by a notable margin on CVC-300. Similarly, SAMed equipped with the proposed strategy attains 92.96\% and 89.42\% on Kvasir-SEG, as well as 94.75\% and 90.54\% on CVC-300, surpassing Polyp-PVT in mDice on both benchmarks.

Table~\ref{tab:synapse_main} presents quantitative results on the Synapse multi-organ CT dataset. For SAMed + our method, the overall mDice reaches 84.95\%, $\uparrow$11.54\% compared with the original SAMed. While nnU-Net and nnFormer achieve slightly higher overall scores, the advantages on small and structurally complex organs are more evident. In particular, segmentation performance on the pancreas, gallbladder, and stomach reaches 92.81\%, 96.74\%, and 89.37\%, corresponding to substantial improvements over nnU-Net ($\uparrow$9.89\%, $\uparrow$25.03\%, and $\uparrow$10.36\%) and nnFormer ($\uparrow$9.46\%, $\uparrow$26.57\%, and $\uparrow$2.54\%). Performance on both kidneys is also notably improved, with left kidney mDice reaching 91.85\% ($\uparrow$5.78\% over nnU-Net, $\uparrow$5.28\% over nnFormer) and right kidney mDice at 90.37\% ($\uparrow$4.12\% over nnFormer). Gains on small and complex organs are particularly pronounced: stomach $\uparrow$21.36\%, pancreas $\uparrow$17.48\%, and gallbladder $\uparrow$13.85\%, far exceeding other SAM-based methods. Specifically, SAMed + our method outperforms existing SOTA approaches on key organs, with gallbladder segmentation $\uparrow$23.48\% over PG-SAM, pancreas $\uparrow$20.28\% over H-SAM, left kidney $\uparrow$3.92\% over PG-SAM, and right kidney $\uparrow$4.38\% over H-SAM, demonstrating the significant advantages of the proposed strategy for segmenting small and complex structures.

\renewcommand{\gain}[1]{\textcolor{gray!60}{\DiffUp\kern0.15em\scalebox{0.92}{#1}}}
\renewcommand{\loss}[1]{\textcolor{gray!60}{\DiffDown\kern0.15em\scalebox{0.92}{#1}}}
\begin{table*}[!t]
\centering
\small 
\caption{Quantitative comparison on the Synapse multi-organ CT dataset. Color coding follows the same convention as Table~\ref{tab:polyp_bench}. mDice: the mean Dice coefficient; HD95: the 95th percentile of the Hausdorff Distance.}
\label{tab:synapse_main}

\setlength{\tabcolsep}{4pt}

\renewcommand{\arraystretch}{1.18}

\resizebox{1.0\textwidth}{!}{
\begin{tabular}{c|c|cccccccc|cc}
\toprule

\textbf{Group} & \textbf{\cellcolor{lightpink1}Method} &
\textbf{\cellcolor{lightpink2}Spleen} & \textbf{\cellcolor{lightpink2}Kidney (R)} & \textbf{\cellcolor{lightpink2}Kidney (L)} &
\textbf{\cellcolor{lightpink2}Gallbladder} & \textbf{\cellcolor{lightpink2}Liver} & \textbf{\cellcolor{lightpink2}Stomach} &
\textbf{\cellcolor{lightpink2}Aorta} & \textbf{\cellcolor{lightpink2}Pancreas} &
\textbf{\cellcolor{lightpink1}mDice[\%]}$\uparrow$ & \textbf{\cellcolor{lightpink1}HD95[mm]}$\downarrow$ \\
\midrule

\multirow{5}{*}{\rotatebox[origin=c]{90}{\centering\bfseries\shortstack{Non\\SAM}}}
& TransUNet~\cite{TransUNet2021}
& 85.08 & 77.02 & 81.87 & 63.16 & 94.08 & 75.62 & 87.23 & 55.86 & 77.48 & 31.69\\
& SwinUNet~\cite{SwinUNet2021}
& 90.66 & 79.61 & 83.28 & 66.53 & 94.29 & 76.60 & 85.47 & 56.58 & 79.13 & 21.55\\
& UNETR~\cite{UNETR2021}
& 87.81 & 84.80 & 85.66 & 60.56 & 94.46 & 73.99 & 89.99 & 59.25 & 79.56 & 18.59 \\
& nnU-Net~\cite{nnUNet2018}
& 90.31 & 91.46 & 86.07 & 71.71 & 95.84 & 79.01 & 92.39 & 82.92 & 86.21 & 10.63 \\
& nnFormer~\cite{nnFormer2022}
& 90.51 & 86.25 & 86.57 & 70.17 & 96.84 & 86.83 & 92.40 & 83.35 & 86.57 & 10.78 \\
\midrule

\multirow{12}{*}{\rotatebox[origin=c]{90}{\centering\bfseries\shortstack{SAM\\based}}}
& SAM3D~\cite{SAM3D2024}
& 84.29 & 85.64 & 86.31 & 49.81 & 95.42 & 76.11 & 89.57 & 69.32 & 79.56 & 23.10\\
& SEG-SAM~\cite{SEGSAM2024}
& 92.10 & 84.16 & 86.39 & 70.21 & 94.91 & 86.06 & 87.42 & 71.48 & 84.09 & 10.26 \\
& H-SAM~\cite{HSAM2024}
& 92.34 & 85.99 & 87.71 & 69.65 & 95.20 & 86.27 & 87.53 & 72.53 & 84.65 & 7.29 \\
& PG-SAM~\cite{PGSAM2025}
& 93.12 & 84.57 & 87.93 & 73.26 & 95.40 & 86.62 & 93.12 & 71.49 & 84.79 & 7.61 \\
\cmidrule(l){2-12} 
& MedSAM~\cite{MedSAM2023}
& 40.61 & 69.29 & 64.36 & 60.17 & 62.69 & 64.67 & 74.69 & 40.32 & 61.31 & 20.37 \\
& \cellcolor{gray!20}\textbf{MedSAM + Ours}
  & \cellcolor{gray!20}\textbf{44.49}\ourscell\,\gain{3.88}
  & \cellcolor{gray!20}\textbf{74.88}\ourscell\,\gain{\textcolor{purple}{5.59}}
  & \cellcolor{gray!20}\textbf{72.08}\ourscell\,\gain{\textcolor{purple}{7.72}}
  & \cellcolor{gray!20}\textbf{64.63}\ourscell\,\gain{4.46}
  & \cellcolor{gray!20}\textbf{65.84}\ourscell\,\gain{3.15}
  & \cellcolor{gray!20}\textbf{67.85}\ourscell\,\gain{3.18}
  & \cellcolor{gray!20}\textbf{80.19}\ourscell\,\gain{\textcolor{purple}{5.50}}
  & \cellcolor{gray!20}\textbf{44.79}\ourscell\,\gain{4.47}
  & \cellcolor{gray!20}\textbf{66.00}\ourscell\,\gain{4.69}
  & \cellcolor{gray!20}\textbf{17.77}\ourscell\,\loss{2.60} \\
& SAM Adapter~\cite{SAMAdapter2023ICCVW}
& 58.23 & 62.07 & 64.57 & 66.89 & 62.54 & 71.10 & 72.34 & 55.72 & 64.18 & 33.08 \\
& \cellcolor{gray!20}\textbf{SAM Adapter + Ours}
  & \cellcolor{gray!20}\textbf{59.12}\ourscell\,\gain{0.89}
  & \cellcolor{gray!20}\textbf{67.23}\ourscell\,\gain{\textcolor{purple}{5.16}}
  & \cellcolor{gray!20}\textbf{68.84}\ourscell\,\gain{4.27}
  & \cellcolor{gray!20}\textbf{71.23}\ourscell\,\gain{4.34}
  & \cellcolor{gray!20}\textbf{64.32}\ourscell\,\gain{1.78}
  & \cellcolor{gray!20}\textbf{75.21}\ourscell\,\gain{4.11}
  & \cellcolor{gray!20}\textbf{75.14}\ourscell\,\gain{2.80}
  & \cellcolor{gray!20}\textbf{60.03}\ourscell\,\gain{4.31}
  & \cellcolor{gray!20}\textbf{67.64}\ourscell\,\gain{3.46}
  & \cellcolor{gray!20}\textbf{17.75}\ourscell\,\loss{\textcolor{purple}{15.33}} \\
& SAMed~\cite{SAMed2024}
& 38.17 & 83.41 & 80.09 & 82.89 & 83.06 & 68.01 & 76.32 & 75.33 & 73.41 & 28.89\\
& \cellcolor{gray!20}\textbf{SAMed + Ours}
  & \cellcolor{gray!20}\textbf{42.20}\ourscell\,\gain{4.03}
  & \cellcolor{gray!20}\textbf{\textcolor{red}{90.37}}\ourscell\,\gain{\textcolor{purple}{6.96}}
  & \cellcolor{gray!20}\textbf{\textcolor{red}{91.85}}\ourscell\,\gain{\textcolor{purple}{11.76}}
  & \cellcolor{gray!20}\textbf{\textcolor{red}{96.74}}\ourscell\,\gain{\textcolor{purple}{13.85}}
  & \cellcolor{gray!20}\textbf{83.12}\ourscell\,\gain{0.06}
  & \cellcolor{gray!20}\textbf{\textcolor{red}{89.37}}\ourscell\,\gain{\textcolor{purple}{21.36}}
  & \cellcolor{gray!20}\textbf{78.10}\ourscell\,\gain{1.78}
  & \cellcolor{gray!20}\textbf{\textcolor{red}{92.81}}\ourscell\,\gain{\textcolor{purple}{17.48}}
  & \cellcolor{gray!20}\textbf{84.95}\ourscell\,\gain{\textcolor{purple}{11.54}}
  & \cellcolor{gray!20}\textbf{\textcolor{red}{7.11}}\ourscell\,\loss{\textcolor{purple}{21.78}} \\
\bottomrule
\end{tabular}
}
\end{table*}

\section{Ablation Studies}

\noindent\textbf{Component-wise Inference Strategy.} We perform a component-wise ablation on Synapse with SAMed as the base model and report mDice (Table~\ref{tab:component_ablation}). The vanilla inference yields 73.41 mDice. Introducing multi-scale candidate proposals improves performance to 79.53 mDice, indicating that prompt diversification is critical under box uncertainty. Incorporating SC scoring further increases mDice to 82.26 by prioritizing stable candidates. Finally, same-$\tau$ top-$n$ fusion achieves the best result of 84.95 mDice, suggesting that aggregating multiple high-confidence hypotheses provides additional robustness beyond ranking. Collectively, the monotonic gains corroborate that candidate generation, stability-aware scoring, and fusion offer complementary contributions to reliable inference.

\begin{center}

  \setlength{\tabcolsep}{0pt}
  \setlength{\fboxsep}{0pt}
  \setlength{\parskip}{0pt}

  \begin{minipage}[t]{0.47\textwidth}
    \centering

    \adjustbox{valign=t, width=\textwidth, keepaspectratio}{%
      \includegraphics{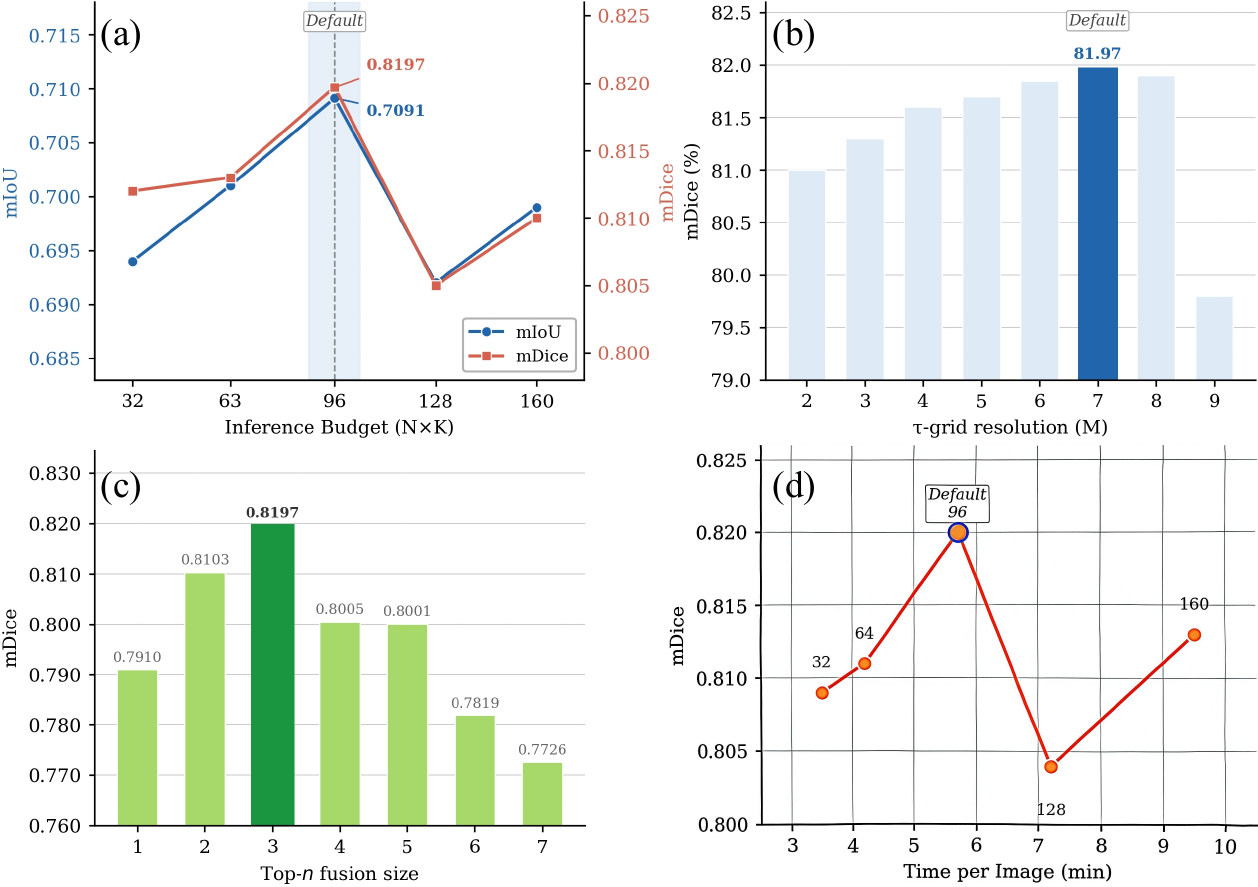}
    }
    \vspace{0.5em}
    \captionof{figure}{Hyper-parameter sensitivity and efficiency characterization on CVC-300 using MedSAM as base model.
    (a) Effect of inference budget on accuracy.
    (b) Sensitivity to $\tau$-grid resolution.
    (c) Sensitivity to top-n fusion size.
    (d) Deploy-time speed–accuracy trade-off.}
    \label{fig:hyperparam_efficiency}
  \end{minipage}
  \hspace{0.03\textwidth}
  \begin{minipage}[t]{0.46\textwidth}
    \centering
    \vspace{0pt} 

    \setlength{\tabcolsep}{1pt}
    \renewcommand{\arraystretch}{1.05}
    \footnotesize  
    \begin{tabular*}{\textwidth}{@{\extracolsep{\fill}} c c c c c @{}}
      \toprule
      \rowcolor{gray!10} 
      \textbf{ID} & \textbf{Can.} & \textbf{SC} & \textbf{Top-$n$} & \textbf{mDice (\%)} \\
      \midrule
      I   &               &             &                 & 73.41 \\
      II  & \ding{51}     &             &                 & 79.53 \\
      III & \ding{51}     & \ding{51}   &                 & 82.26 \\
      IV  & \ding{51}     & \ding{51}   & \ding{51}       & \textbf{84.95} \\
      \bottomrule
    \end{tabular*}
    \vspace{0.5em}

    \captionof{table}{ Component-wise analysis on Synapse with SAMed as the base model. Can. indicates multi-scale box proposals with outer/inner perturbations to generate prompts; SC denotes stability-consistency scoring that ranks candidates by agreement across inner jitters under an image-adaptive threshold set; Top-$n$ denotes same-$\tau$ top-$n$ fusion of the $n$ highest-scoring candidates in probability space before binarization.}
    \label{tab:component_ablation}
  \end{minipage}
\end{center}

\noindent\textbf{Hyper-parameter sensitivity and efficiency characterization.}
We investigate the sensitivity of hyper-parameters on CVC-300 using MedSAM equipped with our inference-time plugin. In Fig.~\ref{fig:hyperparam_efficiency}(a), the budget $\mathrm{Budget}=N{\times}K$ is varied with $M=7$ and $\mathrm{topn}=3$, where performance gains plateau as the budget increases, leading to our default choice of $\mathrm{Budget}=96$. Fig.~\ref{fig:hyperparam_efficiency}(b) examines the effect of threshold-grid resolution $M$ under $N=12$, $K=8$, and $\mathrm{topn}=3$, revealing that $M=7$ offers an optimal balance between accuracy and stability. Fig.~\ref{fig:hyperparam_efficiency}(c) assesses fusion size by varying $\mathrm{topn}$ under default settings, with peak performance observed at a moderate value, confirming $\mathrm{topn}=3$ as a robust choice. Finally, Fig.~\ref{fig:hyperparam_efficiency}(d) illustrates the deploy-time speed--accuracy Pareto frontier for $M=7$ and $\mathrm{topn}=3$, where the operates near the knee point.

\section{Conclusion}
This work presents SAIF, a training-free inference framework that explicitly models joint prompt–threshold uncertainty for box-prompted medical image segmentation with SAM. The proposed stability–consistency scoring mechanism significantly improves prediction reliability and segmentation accuracy across diverse anatomical structures and cross-distribution settings, achieving state-of-the-art performance without any architectural modification or additional training. This is anticipated to inspire further research into inference-time reliability as a fundamental design consideration, thereby advancing the trustworthy deployment of foundation models in safety-critical medical imaging applications.


\newpage
\begin{thebibliography}{22}


\bibitem{survey}
Wang, R., Lei, T., Cui, R., Zhang, B., Meng, H., Nandi, A.K.:
Medical image segmentation using deep learning: A survey.
\textit{IET Image Processing} \textbf{16}(5), 1243--1267 (2022).
\doi{10.1049/ipr2.12419}

\bibitem{SAM2023}
Kirillov, A., Mintun, E., Ravi, N., Mao, H., Rolland, C., Gustafson, L., Xiao, T., Whitehead, S., Berg, A.C., Lo, W.Y., et al.:
Segment Anything.
arXiv preprint arXiv:2304.02643 (2023).
\doi{10.48550/arXiv.2304.02643}

\bibitem{MedSAM2023}
Ma, J., Wang, Y.:
Segment Anything in Medical Images.
\textit{Nature Communications} \textbf{15}(1), 654 (2024).
\doi{10.1038/s41467-024-44824-z}

\bibitem{SAMed2024}
Zhang, K., Liu, D.:
Customized Segment Anything Model for Medical Image Segmentation.
arXiv preprint arXiv:2304.13785 (2023).
\doi{10.48550/arXiv.2304.13785}

\bibitem{SAMAdapter2023ICCVW}
Chen, T., Zhu, L., Zhang, S., Li, Z., Ding, C., Cao, R., Wang, Y., Sun, L., Zang, Y., Mao, P.:
SAM-Adapter: Adapting Segment Anything in Underperformed Scenes.
In: Proceedings of the IEEE/CVF International Conference on Computer Vision Workshops (ICCVW), pp. 3367--3375 (2023).

\bibitem{PGSAM2025}
Zhong, Y., Luo, Z., Liu, C., Tang, F., Hu, Y., Peng, Z., Hu, M., Su, J., Ge, Z., Razzak, I.:
PG-SAM: A Fine-Grained Prior-Guided SAM Framework for Prompt-Free Medical Image Segmentation.
In: 2025 IEEE International Conference on Bioinformatics and Biomedicine (BIBM), pp. 3369–3376. IEEE (2025).
\doi{10.1109/BIBM66473.2025.11356599}

\bibitem{HSAM2024}
Cheng, Z., Wei, Q., Zhu, H., Wang, Y., Qu, L., Shao, W., Zhou, Y.:
Unleashing the Potential of SAM for Medical Adaptation via Hierarchical Decoding.
arXiv preprint arXiv:2403.18271 (2024).
\doi{10.48550/arXiv.2403.18271}

\bibitem{Tang2024DACL}
Tang, F., Xu, Z., Hu, M., Li, W., Xia, P., Zhong, Y., Wu, H., Su, J., Ge, Z.:
Neighbor Does Matter: Density-Aware Contrastive Learning for Medical Semi-supervised Segmentation.
In: AAAI Conference on Artificial Intelligence (AAAI), vol. 39, no. 7, pp. 7449--7457 (2025).
\doi{10.1609/aaai.v39i7.32776}

\bibitem{SAMUS2024}
Lin, X., Xiang, Y., Yu, L., Yan, Z.:
Beyond Adapting SAM: Towards End-to-End Ultrasound Image Segmentation via Auto Prompting.
arXiv preprint arXiv:2309.06824 (2024).
\doi{10.48550/arXiv.2309.06824}


\bibitem{Huang2024MedSAMEval}
Huang, Y., Yang, X., Liu, L., Zhou, H., Chang, A., Zhou, X., Chen, R., Yu, J., Chen, J., Chen, C., et al.:
Segment anything model for medical images?
\textit{Medical Image Analysis} \textbf{92}, 103061 (2024).
\doi{10.1016/j.media.2023.103061}

\bibitem{Synapse2015}
Landman, B., et al.:
Multi-Atlas Labeling Beyond the Cranial Vault.
In: MICCAI Workshop on Multi-Atlas Labeling (MAL), pp. 1--8 (2015).
\doi{10.7303/syn3193805}

\bibitem{CVCClinicDB2015}
Bernal, J., Sánchez, F.J., Fernández-Esparrach, G., Gil, D., Rodríguez, C., Vilariño, F.:
WM-DOVA maps for accurate polyp highlighting in colonoscopy: Validation vs. saliency maps from physicians.
\textit{Computers in Biology and Medicine} \textbf{43}, 99--111 (2015).
\doi{10.1016/j.compmedimag.2015.02.007}

\bibitem{KvasirSEG2020}
Jha, D., Smedsrud, P.H., Riegler, M.A., Halvorsen, P., de Lange, T., Johansen, D., Johansen, H.D.:
Kvasir-SEG: A Segmented Polyp Dataset and Benchmark.
In: International Conference on MultiMedia Modeling (MMM), LNCS, vol. 11962, pp. 451--462. Springer (2020).
\doi{10.1007/978-3-030-37734-2_37}

\bibitem{Vazquez2017}
Vázquez, D., Bernal, J., Sánchez, F.J., Fernández-Esparrach, G., López, A.M., Romero, A., Drozdzal, M., Courville, A.:
A benchmark for endoluminal scene segmentation of colonoscopy images.
\textit{Journal of Healthcare Engineering} \textbf{2017}, 1--10 (2017).

\bibitem{UNet2015}
Ronneberger, O., Fischer, P., Brox, T.:
U-Net: Convolutional Networks for Biomedical Image Segmentation.
In: International Conference on Medical Image Computing and Computer Assisted Intervention (MICCAI), LNCS, vol. 9351, pp. 234--241. Springer, Cham (2015).
\doi{10.1007/978-3-319-24574-4_28}

\bibitem{UNetPlus2018}
Zhou, Z., Siddiquee, M.M.R., Tajbakhsh, N., Liang, J.:
UNet++: A nested U-Net architecture for medical image segmentation.
In: Deep Learning in Medical Image Analysis and Multimodal Learning for Clinical Decision Support, LNCS, vol. 11045, pp. 3--11. Springer, Cham (2018).
\doi{10.1007/978-3-030-00889-5_1}

\bibitem{TransUNet2021}
Chen, J., Lu, Y., Yu, Q., Luo, X., Adeli, E., Wang, Y., Lu, L., Yuille, A.L., Zhou, Y.:
TransUNet: Transformers Make Strong Encoders for Medical Image Segmentation.
arXiv preprint arXiv:2102.04306 (2021). 
\doi{10.48550/arXiv.2102.04306}

\bibitem{EUNet2021}
Patel, K., Bur, A.M., Wang, G.:
Enhanced U-Net: A Feature Enhancement Network for Polyp Segmentation.
In: 2021 18th Conference on Robots and Vision (CRV), pp. 181--188 (2021).
\doi{10.1109/CRV52889.2021.00032}

\bibitem{PraNet2020}
Fan, D.P., Ji, G.P., Zhou, T., Chen, G., Fu, H., Shen, J., Shao, L.:
PraNet: Parallel Reverse Attention Network for Polyp Segmentation.
In: International Conference on Medical Image Computing and Computer-Assisted Intervention (MICCAI), LNCS, vol. 12266, pp. 263--273. Springer, Cham (2020).
\doi{10.1007/978-3-030-59725-2_26}

\bibitem{SwinUNet2021}
Cao, H., Wang, Y., Chen, J., Jiang, D., Zhang, X., Tian, Q., Wang, M.:
Swin-Unet: Unet-like Pure Transformer for Medical Image Segmentation.
In: European Conference on Computer Vision Workshops (ECCVW), LNCS, vol. 13803, pp. 205--218. Springer, Cham (2022).
\doi{10.1007/978-3-031-25066-8_9}

\bibitem{MSNet2021}
Zhao, X., Zhang, L., Lu, H.:
Automatic Polyp Segmentation via Multi-scale Subtraction Network.
In: International Conference on Medical Image Computing and Computer-Assisted Intervention (MICCAI), LNCS, vol. 12901, pp. 120--130. Springer, Cham (2021).
\doi{10.1007/978-3-030-87193-2_12}

\bibitem{PolypPVT2022}
Dong, B., Wang, W., Fan, D.P., Li, J., Fu, H., Shao, L.:
Polyp-PVT: Polyp Segmentation with Pyramid Vision Transformers.
\textit{CAAI Artificial Intelligence Research} \textbf{2}, 1--14 (2023).
\doi{10.26599/AIR.2023.9150015}

\bibitem{UNETR2021}
Hatamizadeh, A., Tang, Y., Nath, V., Yang, D., Myronenko, A., Landman, B., Roth, H.R., Xu, D.:
UNETR: Transformers for 3D Medical Image Segmentation.
In: IEEE/CVF Winter Conference on Applications of Computer Vision (WACV), pp. 574--584 (2022).
\doi{10.1109/WACV51458.2022.00181}

\bibitem{nnUNet2018}
Isensee, F., Jaeger, P.F., Kohl, S.A., Petersen, J., Maier-Hein, K.H.:
nnU-Net: A Self-configuring Method for Deep Learning-based Biomedical Image Segmentation.
\textit{Nature Methods} \textbf{18}(2), 203--211 (2021).
\doi{10.1038/s41592-020-01008-z}

\bibitem{nnFormer2022}
Zhou, H.Y., Guo, J., Zhang, Y., Yu, L., Wang, L., Yu, Y.:
nnFormer: Volumetric Medical Image Segmentation via a 3D Transformer.
\textit{IEEE Transactions on Image Processing} \textbf{32}, 4036--4045 (2023).
\doi{10.1109/TIP.2023.3293771}

\bibitem{SAM3D2024}
Bui, N.T., Hoang, D.H., Tran, M.T., Doretto, G., Adjeroh, D., Patel, B., Choudhary, A., Le, N.:
SAM3D: Segment Anything Model in Volumetric Medical Images.
In: IEEE International Symposium on Biomedical Imaging (ISBI), pp. 1--5 (2024).
\doi{10.1109/ISBI56570.2024.10635844}

\bibitem{SEGSAM2024}
Huang, S., Liang, H., Wang, Q., Zhong, C., Zhou, Z., Shi, M.:
SEG-SAM: Semantic-Guided SAM for Unified Medical Image Segmentation.
arXiv preprint arXiv:2412.12660 (2024).
\doi{10.48550/arXiv.2412.12660}
\end{thebibliography}
\end{document}